\documentclass[11pt,letterpaper]{article}
\usepackage[letterpaper]{geometry}
\usepackage{acl2012}
\usepackage{times}
\usepackage{latexsym}
\usepackage{amsmath,amssymb}
\usepackage{multirow}
\usepackage{url}
\makeatletter
\newcommand{\@BIBLABEL}{\@emptybiblabel}
\newcommand{\@emptybiblabel}[1]{}
\makeatother
\usepackage[draft]{hyperref}

\usepackage{amsmath,amsthm}
\newtheorem*{rep@theorem}{\rep@title}

\usepackage{graphicx,subfigure,color,tikz,multirow,rotating,caption}

\usepackage{algorithm}
\usepackage[noend]{algorithmic}
\def\shownotes{0}  %set 1 to show author notes
\ifnum\shownotes=1
\newcommand{\authnote}[2]{{$\ll$\textsf{\footnotesize #1 notes: #2}$\gg$}}
\else
\newcommand{\authnote}[2]{}
\fi

\newcommand{\yingyu}[1]{{\color{blue}\authnote{Yingyu}{{#1}}}}

\newtheorem{thm}{Theorem}

\newcommand{\inner}[1]{\langle #1\rangle}

\title{Linear Algebraic Structure of Word Senses, with Applications to Polysemy} 

\author{Sanjeev Arora, Yuanzhi Li, Yingyu Liang, Tengyu Ma, Andrej Risteski \\
  Computer Science Department, Princeton University\\
35 Olden St, Princeton, NJ 08540\\
  {\tt \{arora,yuanzhil,yingyul,tengyu,risteski\}@cs.princeton.edu} \\}
	
\date{}

\begin{document}
\maketitle

\begin{abstract}
Word embeddings are ubiquitous in NLP and information retrieval, but it is unclear what they represent when the word is polysemous. 
Here it is shown that multiple word senses reside in linear superposition {\em within} the word embedding and simple sparse coding can recover vectors that approximately capture the senses. 
The success of our approach, which applies to several embedding methods, is mathematically explained using a variant of the {\em random walk on discourses} model~\cite{randomdiscourses}. 
 A novel aspect of our technique is that each extracted word sense is accompanied by one of about $2000$ \textquotedblleft discourse atoms\textquotedblright\ that gives a succinct description of which other words co-occur with that word sense. Discourse atoms can be of independent interest, and make the method potentially more useful.
 Empirical tests are used to verify and support the theory.

% Text version: 
% Word embeddings are ubiquitous in NLP and information retrieval, but it is unclear what they represent when the word is polysemous. Here it is shown that multiple word senses reside in linear superposition within the word embedding and simple sparse coding can recover vectors that approximately capture the senses. The success of our approach, which applies to several embedding methods, is mathematically explained using a variant of the random walk on discourses model (Arora et al., 2016). A novel aspect of our technique is that each extracted word sense is accompanied by one of about 2000 "discourse atoms" that gives a succinct description of which other words co-occur with that word sense. Discourse atoms can be of independent interest, and make the method potentially more useful. Empirical tests are used to verify and support the theory.
\end{abstract}
\section{Introduction} \label{sec:intro}

{\em Word embeddings} 
%represent the \textquotedblleft meaning\textquotedblright\ of a word as a real-valued vector. Their 
are constructed using Firth's hypothesis that a word's sense is captured by the distribution of other words around it~\cite{firth1957a}. Classical vector space models (see the survey by~\newcite{turney2010frequency}) use simple linear algebra on the matrix of word-word co-occurrence counts, whereas recent neural network and energy-based models such as { word2vec} use an objective that involves a nonconvex (thus, also nonlinear) function of the word co-occurrences~\cite{bengio2003neural,mikolov2013distributed,mikolov2013linguistic}. 
%\Tnote{Maybe only 'non-convex'? Non-linear is a subcase of non-convex }

%Word embeddings are useful in many NLP tasks, and seem useful for understanding neural encoding of semantics~\cite{mitchelletal}.

This nonlinearity makes it hard to discern how  these modern embeddings capture the different senses of a polysemous word. The monolithic view of embeddings,  with the internal information extracted only via inner product, is felt to fail in capturing word senses~\cite{griffiths2007topics,reisingermooney,iacobacci2015sensembed}. Researchers have instead sought to capture polysemy using more complicated representations, e.g., by inducing separate embeddings for each sense~\cite{murphytm12learning,huang2012improving}.
These  embedding-per-sense representations grow naturally out of classic  Word Sense Induction or WSI~\cite{yarowsky1995unsupervised,schutze98,reisingermooney,di2013clustering} techniques that perform clustering on neighboring words. 

The current paper goes beyond this monolithic view, by describing how multiple senses of a word actually reside in linear superposition {\em within} the standard word embeddings (e.g., word2vec~\cite{mikolov2013distributed} and GloVe~\cite{pennington2014glove}). By this we mean the following: consider a polysemous word, say {\em tie}, which can refer to an article of clothing, or a drawn match, or a physical act. Let's take the usual viewpoint that {\em tie} is a single token that represents monosemous words {\em tie1, tie2, ...}.
%Let's assume these canonical senses of {\em tie} occur with very different distributions of neighboring words in the corpus.
%\footnote{Equating word senses with distributions of
%neighboring words leads to well-known controversies in lexicography, e.g., it implies different senses of {\em paint}  in {\em paint the wall} versus {\em paint a mural}. }
% where \textquotedblleft unrelated\textquotedblright\ means that 
\iffalse Similarly, the word distribution around the \textquotedblleft article of clothing\textquotedblright\ sense of {\em tie} is very different from that around the \textquotedblleft physical act\textquotedblright\ sense even though the two are clearly metaphorically related.}\fi 
%The main proposal in this paper, supported by theory and supporting experiments, is that for the standard word embeddings (where the vector need not be unit norm):
The theory and experiments in this paper strongly suggest that word embeddings computed using modern techniques such as GloVe and word2vec satisfy:
%We suggest ---using theory and experiments---that
\vspace{-0.02in}
\begin{align} \label{eqn:tieexpansion}
	v_{\text{\em tie}} \approx \alpha_1 \, v_{\text{\em tie1}} + \alpha_2 \, v_{\text{\em tie2}} + \alpha_3  \, v_{\text{\em tie3}} +\cdots
\end{align}
where coefficients $\alpha_i$'s are nonnegative and $v_{tie1}, v_{tie2},$ etc., are the hypothetical embeddings of the different senses---those that {\em would} have been induced in the thought experiment where
all occurrences of the different senses were hand-labeled in the corpus.
This {\em Linearity Assertion}, whereby linear structure appears out of a highly nonlinear embedding technique,  is explained theoretically in Section~\ref{sec:linearitythm}, and then empirically tested in a couple of ways in Section~\ref{sec:explinear}.

%{\em A priori} it seems unclear how to verify the above suggestion.
% full experimental confirmation of this insight % verifying this conjecture/insight 
 % gives a  theoretical explanation for the phenomenon.
% which  also ends up revealing (see Section~\ref{subsec:overlap}) a close relationship between modern word embeddings and the 
%above-mentioned embedding-per-sense representations.  

 Section~\ref{sec:method} uses the linearity assertion to  show how to do WSI via sparse coding, which can be seen as a linear algebraic analog of the classic clustering-based approaches, albeit with overlapping clusters. On  standard testbeds it is competitive with earlier embedding-for-each-sense approaches (Section~\ref{sec:test}).  A novelty of our WSI method is that it automatically links different senses of different words via our {\em atoms of discourse} (Section~\ref{sec:method}). 
 This can be seen as an answer to the suggestion in~\cite{reisingermooney} to enhance one-embedding-per-sense methods so that they can automatically link together senses for different words, e.g., recognize that the \textquotedblleft article of clothing\textquotedblright\   sense of {\em tie} is connected to {\em shoe, jacket,} etc.
%In particular, it also produces canonical {\em discourse} vectors that connect 

\iffalse 
Move later

An important novel aspect of our method is that senses of different words are interconnected via the notion of {\em atoms of discourse}, a new notion introduced in Section~\ref{sec:method}. To give an example, the \textquotedblleft article of clothing\textquotedblright\ sense of {\em tie} is automatically related to words like {\em trousers, blouse,} etc.\ (see Table~\ref{tab:representation}). This makes it potentially more useful for some applications.
%, such as automated construction of WordNets in foreign languages. 

Section~\ref{sec:exp} describes the experimental results on atoms of discourse, and Section~\ref{sec:test} describes the performance of our WSI method and compares it to the performance of humans as well as traditional graph clustering-based approaches. 

\fi 

This paper is inspired by the solution of word analogies via linear algebraic methods~\cite{mikolov2013linguistic}, and use of sparse coding on word embeddings to get useful representations  for many NLP tasks~\cite{faruqui2015sparse}. Our theory builds conceptually upon the {\em random walk on discourses} model of~\newcite{randomdiscourses}, although we make a small but important change  to explain empirical findings regarding polysemy.
  Our WSI procedure applies (with minor variation in performance) to canonical embeddings such as {word2vec} and {GloVe} as well as the older vector space methods such as PMI~\cite{church1990word}.
 % The basic insights from the mathematical explanation in Section~\ref{subsec:explanation} carry over, since these embedding 
 This is not surprising since these embeddings are known to be interrelated~\cite{levy2014neural,randomdiscourses}.

%We present in Sections~\ref{??} experimental results leveraging our ideas in other settings where WSI procedures have been tested.

\section{Justification for Linearity Assertion}
\label{sec:linearitythm}

Since word embeddings are solutions to nonconvex optimization problems, at first sight it appears hopeless  to reason about their finer structure. But it becomes possible to do so using a generative model for language~\cite{randomdiscourses} --- a dynamic versions by the log-linear topic model of~\cite{mnih2007three}---which we now recall.
It posits that at every point in the corpus there is a micro-topic (\textquotedblleft what is being talked about\textquotedblright) called {\em discourse} that is drawn from the continuum of unit vectors in $\Re^d$. The parameters of the model include  a vector $v_w \in \Re^d$ for each word $w$. Each discourse $c$ defines a distribution over words 
$
   \Pr[w~|~c] \propto \exp(c \cdot v_w).
 $
 The model assumes that the corpus is generated by the slow geometric random walk of $c$ over the unit sphere in $\Re^d$~: when the walk is at  $c$, a few words are emitted by i.i.d.\ samples from the distribution~(\ref{eqn:discoursemodel}), which, due to its log-linear form, strongly favors words close to $c$ in cosine similarity.
 Estimates for learning  parameters $v_w$ using MLE and moment methods correspond to standard embedding methods such as GloVe and word2vec (see the original paper).

 To study how word embeddings capture word senses, we'll need to understand the relationship between a word's embedding and those of words it co-occurs with.  
In the next subsection, we propose a slight modification to the above model and shows how to infer the embedding of a word from the embeddings of other words that co-occur with it. 
This immediately leads to the Linearity Assertion, as shown in Section~\ref{sec:proof_linear}. 

\yingyu{reorganize and rewrite the subsections; due to that the reviewers asking for more explanation  about background and also Thm 1}

\subsection{Gaussian Walk Model} \label{subsec:ellipsoid}

As alluded to before, we modify the random walk model of~\cite{randomdiscourses} to the {\em Gaussian random walk model}. Again, the parameters of the model include a vector $v_w \in \Re^d$ for each word $w$.
The model assumes the corpus is generated as follows. 
First, a discourse vector $c$ is drawn from a Gaussian with mean $0$ and covariance $\Sigma$. Then, a window of $n$ words $w_1, w_2, \dots, w_n$ are generated from $c$ by: 
\begin{align} \label{eqn:discoursemodel}
   \Pr[w_1,w_2,\dots,w_n|~c] & = \prod_{i=1}^n \Pr[w_i|~c], \\
   \Pr[w_i~|~c] & = \exp(c \cdot v_{w_i}) / Z_c,
\end{align}   
where $Z_c = \sum_w \exp(\langle v_w, c\rangle)$ is the partition function. 
We also assume the partition function concentrates in the sense that  $Z_c \approx Z \exp(\|c\|^2)$ for some constant $Z$.  This is a direct extension of~\cite[Lemma 2.1]{randomdiscourses} to discourse vectors with norm other than 1, and causes the additional term $\exp(\|c\|^2)$.\footnote{The formal proof of~\cite{randomdiscourses} still applies in this setting. The simplest way to informally justify this assumption is to assume $v_w$ are random vectors, and then $Z_c$ can be shown to concentrate around $\exp(\|c\|^2)$.  Such a condition enforces the word vectors to be isotropic to some extent, and makes the covariance of the discourse identifiable. } %We refer to this model as the 

%We assume the Gaussian random walk and the partition function is concentrated in the sense above in the following proof of Theorem~\ref{thm:ellipsoid1}.  %We restate a more detailed version of Theorem~\ref{thm:ellipsoid1}.

%\begin{thm}[Formal version of Theorem~\ref{thm:ellipsoid1}] \label{thm:ellipsoid} Assume the Gaussian walk model and the partition function is concentrated the sense above. Then, we have for any fixed word $w$, 
%	$$v_w = A~\mathbb{E}\big[\sum_{w_i \in s} w_i\mid w\in s\big]\nonumber$$ for a matrix $A =  (\Sigma^{-1}+ 2I) (\Sigma^{-1}+ 2nI)^{-1}$. 
%\end{thm} 

%Note that this simple fix also explains why the older polysemy models was still useful in downstream applications: the required %linear transformation may be implicitly learnt downstream (e.g., via a linear classifier or a neural net).
%\Tnote{TODO?: explain why $\Sigma$ can be identifiable.}

\begin{thm} \label{thm:ellipsoid1} 
		Assume the above generative model, and let $s$ denote the random variable of a window of $n$ words.
 Then, there is a linear transformation $A$ such that 
 $v_w \approx  A~\mathbb{E}\left[\frac{1}{n}\sum_{w_i \in s} v_{w_i}\mid w\in s\right]$. 
 \end{thm} 

\begin{proof}
Let $c_s$ be the discourse vector for the whole window $s$. By the law of total expectation, we have
	\begin{align}
	&\mathbb{E}\left[c_s \mid w\in s\right] \nonumber\\
	= &	\mathbb{E}\left[\mathbb{E}[c_s \mid s=w_1\dots w_{j-1}ww_{j+1}\dots w_n]\mid w\in s\right]. \label{eqn:exp1}
	\end{align}
	We evaluate the two sides of the equation. 
	
	First, by Bayes' rule and the assumptions on the distribution of $c$ and the partition function, we have:
	\begin{align*}
	p(c|w)& \propto p(w|c)p(c)\nonumber\\
	&\propto \frac{1}{Z_c} \exp(\langle v_w, c\rangle ) \cdot \exp\left(-\frac 1 2 c^\top \Sigma^{-1}c\right)\nonumber\\
	& \approx \frac{1}{Z} \exp\left(\langle v_w, c\rangle - c^\top \left(\frac 1 2 \Sigma^{-1}+I \right) c\right). \nonumber
	\end{align*}
	So $c\mid w$ is a Gaussian distribution with mean 
	\begin{align}
	\mathbb{E}\left[c\mid w\right] \approx (\Sigma^{-1}+ 2I)^{-1}v_w. \label{eqn:lhs}
	\end{align}
	
	Next, we compute $\mathbb{E}[c|w_1, \dots, w_n]$. Again using Bayes' rule and the assumptions on the distribution of $c$ and the partition function, 
	\begin{align*}
	& \quad p(c|w_1, \dots, w_n) \nonumber \\
	& \propto  p(w_1, \dots, w_n|c) p(c) \nonumber \\
	& \propto p(c)\prod_{i=1}^n p(w_i|c) \nonumber \\
	& \approx \frac{1}{Z^n} \exp \left( \sum_{i=1}^n v_{w_i}^\top c - c^\top \left(\frac 1 2\Sigma^{-1} + nI \right) c\right). %\label{eqn:MAP}
	\end{align*}
	So $c|w_1\dots w_n$ is a Gaussian distribution with mean
	\begin{align}
	\mathbb{E}[c|w_1, \dots, w_n] \approx \left(\Sigma^{-1} + 2nI \right)^{-1} \sum_{i=1}^n v_{w_i}.\label{eqn:rhs}
	\end{align}
	Now plugging in equation~\eqref{eqn:lhs} and~\eqref{eqn:rhs} into equation~\eqref{eqn:exp1}, we conclude that 
	{\small 
	\begin{align}
	 (\Sigma^{-1}+ 2I)^{-1}v_w \approx (\Sigma^{-1} + 2nI )^{-1}\mathbb{E}\left[ \sum_{i=1}^n v_{w_i} \mid w\in s\right].\nonumber
	\end{align}
	}
	Re-arranging the equation completes the proof with $A = n(\Sigma^{-1}+ 2I) (\Sigma^{-1}+ 2nI)^{-1}$. 
\end{proof}

\noindent{\bf Note: Interpretation.} 
\yingyu{reviewers required some  explanation; especially reviewer D} 
Theorem~\ref{thm:ellipsoid1} shows that there exists a linear relationship between the vector of a word and the vectors of the words in its contexts. Consider the following thought experiment. First, choose a word $w$. Then, for each window $s$ containing $w$, take the average of the vectors of the words in $s$ and denote it as $v_s$. Now, take the average of $v_s$ for all the windows $s$ containing $w$, and denote the average as $u$. Theorem~\ref{thm:ellipsoid1} says that $u$ can be mapped to the word vector $v_w$ by a linear transformation that does not depend on $w$.
This linear structure may also have connections to some other phenomena related to linearity, e.g.,~\newcite{gittens2017skip} and \newcite{tian2017mechanism}. Exploring such connections is left for future work.

The linear transformation is closely related to $\Sigma$, which describes the distribution of the discourses.
If we choose a coordinate system such that $\Sigma$ is a diagonal matrix with diagonal entries $\lambda_i$, then $A$ will also be a diagonal matrix with diagonal entries $(n+2n \lambda_i) / (1+2n \lambda_i)$. This is smoothing the spectrum and essentially shrinks the directions corresponding to large $\lambda_i$ relatively to the other directions. Such directions are for common discourses and thus common words.
%This doesn't mean that the transformation matrix $A$ is close to a scaled identity matrix. 
%Empirically, we observed that $A$ shrinks the common components of the word vectors. For example, if $v$ is the first singular vector of the word vectors, then $\|Av\|_2$ is much smaller than $\|v\|_2$. 
Empirically, we indeed observe that $A$ shrinks the directions of common words. For example, its last right singular vector has, as nearest neighbors, the vectors for words like ``with'', ``as'', and ``the.'' 
Note that empirically, $A$ is not a diagonal matrix since the word vectors are not in the coordinate system mentioned.  
\yingyu{address the misunderstanding of reviewer B about $A$}

\noindent{\bf Note: Implications for GloVe and word2vec.}  
Repeating the calculation in~\newcite{randomdiscourses} for our new generative model, we can show that the solutions to GloVe and word2vec training objectives solve for the following vectors:
$
  \hat{v}_w = \left(\Sigma^{-1} + 4 I\right)^{-1/2} v_w.
$
Since these other embeddings are the same as $v_w$'s up to linear transformation, Theorem~\ref{thm:ellipsoid1} (and the Linearity Assertion) still holds for them. 
% by replacing $v_w$'s with the trained vectors $\hat{v}_w$'s.
Empirically, we find that $\left(\Sigma^{-1} + 4 I\right)^{-1/2}$ is close to a scaled identity matrix (since $\|\Sigma^{-1}\|_2$ is small), so $\hat{v}_w$'s can be used as a surrogate of $v_w$'s.

\noindent{\bf Experimental note: Using better sentence embeddings, SIF embeddings.} 
Theorem~\ref{thm:ellipsoid1} implicitly uses the average of the neighboring word vectors as an estimate (MLE) for the discourse vector. This estimate is of course also a simple {\em sentence embedding}, very popular in  empirical NLP work and also reminiscent of word2vec's training objective. In practice, this naive sentence embedding can be improved by taking a weighted combination (often tf-idf) of adjacent words.
 The paper~\cite{arora2016simple}  uses a simple twist to the generative model in~\cite{randomdiscourses} to provide a better estimate of the discourse $c$ called {SIF embedding}, which is better for downstream tasks and surprisingly competitive with sophisticated LSTM-based sentence embeddings. 
 It is a {\em weighted} average of word embeddings in the window, with smaller weights for more frequent words (reminiscent of tf-idf). This weighted average is the MLE estimate of $c$ if  above generative model is changed to: 
 $$p(w|c) = \alpha p(w) + (1-\alpha)\frac{\exp(v_w\cdot c)}{Z_c},$$
where $p(w)$ is the overall probability of word $w$ in the corpus and $\alpha>0$ is a constant (hyperparameter).
 
 The theory in the current paper works with SIF embeddings as an estimate of the discourse $c$; in other words,  in Theorem~\ref{thm:ellipsoid1} we replace the average word vector with the SIF vector of that window. Empirically we find that it leads to similar results in testing our theory (Section~\ref{sec:explinear}) and better results in downstream WSI applications (Section~\ref{sec:test}).  Therefore, SIF embeddings are adopted in our experiments.

\subsection{Proof of Linearity Assertion} \label{sec:proof_linear}

Now we use Theorem~\ref{thm:ellipsoid1} to show how the 
 Linearity Assertion follows. Recall the thought experiment considered there. Suppose word $w$ has two distinct senses $s_1$ and $s_2$. Compute a word embedding $v_w$ for $w$.  Then hand-replace each occurrence of a sense of $w$ by one of the new tokens $s_1, s_2$  depending upon which one is being used. Next, train separate embeddings for $s_1, s_2$ while keeping the other embeddings fixed. (NB: the  classic clustering-based sense induction~\cite{schutze98,reisingermooney} can be seen as an approximation to this thought experiment.) 

\begin{thm}[Main] \label{thm:main1} 
Assuming the model of Section~\ref{subsec:ellipsoid}, embeddings in the thought experiment above will satisfy
$\|v_w - \bar{v}_w\|_2 \rightarrow 0$ as the corpus length tends to infinity, where $\bar{v}_w \approx \alpha v_{s_1} + \beta v_{s_2}$ for
%$$\alpha = \frac{f_1\sqrt{\log(f_1 + f_2)}}{(f_1 + f_2)\sqrt{\log f_1}},\quad \beta = \frac{f_2 \sqrt{\log(f_1 + f_2)}}{(f_1+ f_2)\sqrt{\log f_2}},$$
$$\alpha = \frac{f_1}{f_1 + f_2},\quad \beta = \frac{f_2}{f_1+ f_2},$$
where $f_1$ and $f_2$ are the numbers of occurrences of $s_1, s_2$ in the corpus, respectively. 
\yingyu{corrected $\alpha, \beta$}
\end{thm}

\begin{proof}
%As noted in Section~\ref{sec:pastmodels}, for each occurrence of $w$, we get an estimate for its embeddings, namely, the estimate of the discourse vector $c$ that generated it and the surrounding words. 
%For simplicity we first prove in the simpler random walk (i.e. Lemma~\ref{lem:vestimate}).
Suppose we pick a random sample of $N$ windows containing $w$ in the corpus. %Assuming the random walked mixed between these occurrences,  
For each window, compute the average of the word vectors and then apply the linear transformation in Theorem~\ref{thm:ellipsoid1}. 
The transformed vectors are i.i.d.\ estimates for $v_w$, but with high probability about $f_1/(f_1 + f_2)$ fraction of the occurrences used sense $s_1$ and $f_2/(f_1 +f_2)$ used sense $s_2$, and the corresponding estimates for those two subpopulations converge to $v_{s_1}$ and $v_{s_2}$ respectively. Thus by construction, the estimate for $v_w$ is a linear combination of those for $v_{s_1}$ and $v_{s_2}$. 
%The argument so far establishes linearity but doesn't determine the coefficients in the linear combination. 
%These are determined by a simple calculation using the fact that in the generative model the length of the word embedding is proportional to the square root of the log of its frequency (see Theorem 2.2 in ~\cite{randomdiscourses}).
\end{proof} 

\noindent{\bf Note.} Theorem~\ref{thm:ellipsoid1} (and hence the Linearity Assertion) holds already for the original model in~\newcite{randomdiscourses} but with $A= I,$ where $I$ is the identity transformation. In practice, we find inducing the word vector requires a non-identity $A$, which is the reason for the modified model of Section~\ref{subsec:ellipsoid}.
This also helps to address a nagging issue
hiding in older clustering-based approaches such as~\newcite{reisingermooney} and \newcite{huang2012improving}, which identified senses of a polysemous word by clustering the sentences that contain it.  One imagines a good representation of the sense of an individual cluster is simply the cluster center. 
   This turns out to be false --- the closest words to the cluster center sometimes are not meaningful for the sense that is  being captured; see Table~\ref{tab:huang_center}. Indeed, the authors of~\newcite{reisingermooney} seem aware of this because they mention \textquotedblleft We do not assume that clusters correspond to traditional word senses. Rather, we only rely on clusters to capture meaningful variation in word usage.\textquotedblright\ 
    %However, applying our linear transformation to cluster centers makes them meaningful again. 
		We find that applying $A$ to cluster centers makes them meaningful again. See also Table 1.
    
    \begin{table}[!t]
    	\centering
    	{\footnotesize
    		\begin{tabular}{l|l|l}
    			\hline 
    			\multirow{ 2}{*}{center 1} & before  & and provide providing a \\
    			 & after  & providing provide opportunities provision\\
    			\hline 
    			\multirow{ 2}{*}{center 2} & before  & and a to the \\
    			 & after  & access accessible allowing provide\\
    			\hline 
    		\end{tabular}
    	}
    	\caption{Four  nearest words for some cluster centers that were computed for the word ``access'' by applying $5$-means on the estimated discourse vectors (see Section~\ref{subsec:ellipsoid}) of 1000 random windows from Wikipedia containing ``access''. After applying the linear transformation of Theorem~\ref{thm:ellipsoid1} to the center, the nearest words become meaningful. } 
    	\label{tab:huang_center}
    \end{table}

\section{Towards WSI: Atoms of Discourse} \label{sec:method}

Now we consider how to do WSI using only word embeddings and the Linearity Assertion. Our approach is fully unsupervised, and tries to induce  senses for all words in one go, together with a vector representation for each sense. % that can be linked across 
% This goal does not easily lend itself to simple clustering (eg k-means), in contrast to many classic benchmark test settings (also discussed in Section~\ref{sec:test}), where a single word is given together with its occurences, and the senses can indeed be identified by simple clustering. 

%linear algebraic decomposition of word senses as in (\ref{eqn:tieexpansion}).
% given the 
 \iffalse
 The above experiment and its theoretical explanation suggests that the embedding of a polysemous word like {\em tie} can be rewritten as 
 \begin{align} \label{eqn:tieexpansion}
	v_{\text{\em tie}} \approx \alpha_1 \, v_{\text{\em tie1}} + \alpha_2 \, v_{\text{\em tie2}} + \alpha_3  \, v_{\text{\em tie3}} +\cdots
\end{align}
where $v_{tie1}, v_{tie2},$ etc., are the hypothetical embeddings of the different senses of {\em tie}, and

% $\alpha_i$'s are related to the frequency with which these senses occur in the corpus.
It is unclear how to proceed since in that expression
 $v_{tie1}, v_{tie2},$ etc., and $\alpha_i$'s are unknown.
 \fi 
Given embeddings for all words, it seems unclear at first sight how to pin down the senses of {\em tie} using only (\ref{eqn:tieexpansion})  since
%But it is unclear how expression (\ref{eqn:tieexpansion}) alone could allow us to pin down 
%the senses 
 $v_{tie}$ can be expressed in infinitely many ways as such a combination, and this
is true even if $\alpha_i$'s were known (and they aren't).
% so (\ref{eqn:tieexpansion}) is insufficient to pin down the senses.
%\Tnote{the sentence above might be confusing (as Daichi pointed out). How about ``There the unknowns are the embeddings of the senses of each word and the coefficients before them, while we don't have enough observations (that is, the word vectors) to pin them down. ''}
%{\sc sanjeev: he was confused about whether $\alpha_i$'s were known. This is now clarified}
 To pin down the senses we will need to interrelate the senses of different words, for example, relate the \textquotedblleft article of clothing\textquotedblright\ sense {\em tie1} with  {\em shoe, jacket,} etc. 
 To do so we rely on the generative model of Section~\ref{subsec:ellipsoid} according to which unit vector in the embedding space corresponds to a micro-topic or discourse. Empirically, discourses  $c$ and $c'$ tend to look similar to humans (in terms of nearby words) if their inner product is larger than $0.85$, and quite different if the inner product is smaller than $0.5$. So in the discussion below, a discourse should really be thought of as a  small region rather than a point.
One imagines that the corpus has a \textquotedblleft clothing\textquotedblright\ discourse that has a high probability of outputting the {\em tie1} sense, and also of outputting related words such as {\em shoe, jacket,} etc.
By (\ref{eqn:discoursemodel}) the probability of being output by a discourse is determined by the inner product, so one expects that the vector  for \textquotedblleft clothing\textquotedblright\ discourse  has a high inner product with all of {\em shoe, jacket, tie1,} etc., and thus can stand as surrogate for $v_{\text{\em tie1}}$ in (\ref{eqn:tieexpansion})!
Thus it may be sufficient to consider the following global optimization:
 
{\em Given word vectors $\{v_w\}$ in $\Re^d$ and two integers $k, m$ with $k < m$, find a set of unit vectors $A_1, A_2, \ldots, A_m$ such that
\vspace{-0.05in}\small
\begin{align} \label{eqn:sparsecoding}
  v_w = \sum_{j=1}^m\alpha_{w,j}A_j + \eta_w
\end{align}
\normalsize
where at most $k$ of the coefficients $\alpha_{w,1},\dots,\alpha_{w,m}$ are nonzero, and $\eta_w$'s are error vectors.} 

Here $k$ is the sparsity parameter, and $m$ is the number of atoms, and the optimization minimizes the norms of $\eta_w$'s (the $\ell_2$-reconstruction error): 
\vspace{-.03in}
\small
\begin{align} \label{eqn:sparsecoding_obj}
 \sum_w \bigg\|v_w - \sum_{j=1}^m\alpha_{w,j}A_j \bigg\|_2^2.
\end{align}
\normalsize
Both $A_j$'s and $\alpha_{w,j}$'s are unknowns, and the optimization is nonconvex. This is just {\em sparse coding}, useful in neuroscience~\cite{OF} and also in image processing, computer vision, etc. 

This optimization is a surrogate for the desired expansion of $v_{\text{\em tie}}$ as in (\ref{eqn:tieexpansion}), because one can hope that among $A_1, \ldots, A_m$ there will be directions corresponding to {\em clothing}, {\em sports matches,} etc., that will have high inner products with {\em tie1}, {\em tie2,} etc., respectively. 
Furthermore, restricting $m$ to be much smaller than the number of words ensures that the typical $A_i$ needs to be reused to express multiple words.

We refer to $A_i$'s, discovered by this procedure, as {\em atoms of discourse}, since experimentation suggests that the actual discourse in a typical place in text (namely, vector $c$ in (\ref{eqn:discoursemodel})) is a linear combination of a small number, around 3-4, of such atoms. Implications of this for text analysis are left for future work.

\noindent\textbf{Relationship to Clustering.} \label{subsec:overlap}
Sparse coding 
%as described in (\ref{eqn:sparsecoding}) is usually solved (for any choice of $m, k$) by using 
is solved using alternating minimization to find the $A_i$'s that minimize (\ref{eqn:sparsecoding_obj}).
This objective function reveals sparse coding to be a linear algebraic analogue of {\em overlapping} clustering, whereby the $A_i$'s act as {\em cluster centers} and each $v_w$ is assigned in a soft way to at most $k$ of them (using the coefficients $\alpha_{w,j}$, of which at most $k$ are nonzero). In fact this clustering viewpoint is also the basis of the alternating minimization algorithm. In the special case when $k=1$, each $v_w$ has to be assigned to a single cluster, which is the familiar geometric clustering with squared $\ell_2$ distance.

Similar overlapping clustering in a traditional graph-theoretic setup ---clustering while simultaneously cross-relating the senses of different words---seems more difficult but worth exploring. %\yingyu{maybe there are related work; I'll search}
% for future work.
\vspace{-0.02in}

\section{Experimental Tests of Theory} 
\label{sec:explinear}

\subsection{Test of Gaussian Walk Model: Induced Embeddings} 
\label{sec:ellipsoidtest}
Now we test the prediction of the Gaussian walk model suggesting a linear method to induce embeddings from the context of a word. Start with the GloVe embeddings; let $v_w$ denote the embedding for $w$. Randomly sample many paragraphs from Wikipedia, and for each word $w'$ and each occurrence of $w'$ compute the SIF embedding of text in the window of $20$ words centered around $w'$. 
Average the SIF embeddings for all occurrences of $w'$ to obtain vector $u_{w'}$. The Gaussian walk model says that there is a 
linear transformation that maps $u_{w'}$ to $v_{w'}$, so solve the regression:
\begin{equation} \label{eqn:induced}
\text{argmin}_{A} \sum_w \|A u_{w} - v_w \|_2^2. 
\end{equation}
We call the vectors $A u_{w}$ the {\em induced embeddings.}
We can test this method of inducing embeddings by holding out 1/3 words randomly, doing the regression (\ref{eqn:induced}) on the rest, and computing the cosine similarities between $Au_w$ and $v_w$ on the heldout set of words. 
% (\ref{eqn:2}), and used the transformed average discourse vectors as the induced word vectors.

%Table~\ref{tab:wiki_convergence} 
%shows the result. 
%The cosine similarity increases with the amount of data for computing the discourse vectors and plateaus with about a million sentences. 
Table~\ref{tab:wiki_convergence}  shows that the average cosine similarity between the induced embeddings and the GloVe vectors is large. By contrast the average similarity between the average discourse vectors and the GloVe vectors is much smaller (about 0.58), illustrating the need for the linear transformation. %Lemma~\ref{lem:vestimate}.
Similar results are observed for the word2vec and SN vectors~\cite{randomdiscourses}. 
 
%Our analysis suggests the induced vectors are better at capturing the senses, and this will be verified in the experiments in Section~\ref{sec:test}. 

%\begin{table}[!t]
	%\centering
		%\begin{tabular}{c|c c c c c}
		%\hline			
  %\#paragraphs & 250k & 500k &  750k & 1 million  \\
	%\hline
	%relative error &   0.38 & 0.35 & 0.34 & 0.34 \\
	%cosine  & 0.94 & 0.95 & 0.96 & 0.96 \\
		%\hline
		%\end{tabular}
	%\caption{Relative errors of fitting the GloVe word vectors with average discourse vectors using a linear transformation.
	%Row 1 to Row 3: the number of paragraphs used to compute the discourse vectors, the relative error of the fitting, the average cosine similarities between the fitted vectors and the GloVe vectors.}
	%\label{tab:wiki_convergence}
%\end{table}

\begin{table}[!t]
	\centering
		\begin{tabular}{c|c c c c c}
		\hline			
  \#paragraphs & 250k & 500k &  750k & 1 million  \\
	\hline
	cos similarity  & 0.94 & 0.95 & 0.96 & 0.96 \\
		\hline
		\end{tabular}
	\caption{Fitting the GloVe word vectors with average discourse vectors using a linear transformation.
	The first row is the number of paragraphs used to compute the discourse vectors, and the second row is the average cosine similarities between the fitted vectors and the GloVe vectors.}
	\label{tab:wiki_convergence}
\end{table}

\subsection{Test of Linearity Assertion}

\yingyu{remove the old experiments, add the new pseudoword exp}
We do two empirical tests of the Linearity Assertion (Theorem~\ref{thm:main1}).

\noindent\textbf{Test 1.} The first test involves the classic artificial polysemous words (also called pseudowords).
First, pre-train a set $W_1$ of word vectors on Wikipedia with existing embedding methods. Then, randomly pick $m$ pairs of non-repeated words, and for each pair, replace each occurrence of either of the two words with a pseudoword. 
Third, train a set $W_2$ of vectors on the new corpus, while holding fixed the vectors of words that were not involved in the pseudowords. 
%, and compare the vectors for these pseudowords with those predicted by Theorem~\ref{thm:main1}. 
Construction has ensured that each pseudoword has two distinct ``senses", and we
also have in $W_1$ the ``ground truth" vectors for those senses.\footnote{Note that this discussion assumes that the set of pseudowords is small, so that a typical neighborhood of a pseudoword does not consist of other pseudowords. Otherwise the ground truth vectors in $W_1$ become a bad approximation to the sense vectors.} Theorem~\ref{thm:main1} implies that
the embedding of a pseudoword is a linear combination of the sense vectors, so we can compare this predicted embedding to the one learned in $W_2$.\footnote{Here $W_2$ is trained while fixing the vectors of words not involved in pseudowords to be their pre-trained vectors in $W_1$. We can also train all the vectors in $W_2$ from random initialization. Such $W_2$ will not be aligned with $W_1$. 
Then we can learn a linear transformation from $W_2$ to $W_1$ using the vectors for the words not involved in pseudowords, apply it on the vectors for the pseudowords, and compare the transformed vectors to the predicted ones. This is tested on word2vec, resulting in relative errors between $20\%$ and $32\%$, and cosine similarities between $0.86$ and $0.92$. These results again support our analysis. 
}

\begin{table}[!t]
	\centering
		\begin{tabular}{c| c | c c  c}
			\hline 
			\multicolumn{2}{c}{$m$ pairs}  & $10$ & $10^3$ & $3 \cdot 10^4$\\
			\hline
			\multirow{2}{*}{relative error} & SN  & 0.32 & 0.63 & 0.67\\
			                       &GloVe & 0.29  & 0.32 & 0.51 \\ 
			\hline
			\multirow{2}{*}{cos similarity} & SN  & 0.90 & 0.72 & 0.75\\
			                       &GloVe & 0.91  & 0.91 & 0.77 \\ 
			\hline
		\end{tabular}
	\caption{The average relative errors and cosine similarities between the vectors of pseudowords and those predicted by Theorem~\ref{thm:main1}. $m$ pairs of words are randomly selected and for each pair, all occurrences of the two words in the corpus is replaced by a pseudoword. Then train the vectors for the pseudowords on the new corpus.}
	\label{tab:pseudoword}
\end{table}

Suppose the trained vector for a pseudoword $w$ is $u_{w}$ and the predicted vector is $v_{w}$, then the comparison criterion is the average relative error
$
  \frac{1}{|S|}\sum_{w \in S} \frac{\|u_{w} - v_w\|^2_2 }{\|v_w\|^2_2}
$
where $S$ is the set of all the pseudowords.
We also report the average cosine similarity between $v_w$'s and $u_w$'s.

Table~\ref{tab:pseudoword} shows the results for the GloVe and SN~\cite{randomdiscourses} vectors, averaged over 5 runs. When $m$ is small, the error is small and the cosine similarity is as large as $0.9$. Even if $m = 3\cdot 10^4$ (i.e., about $90\%$ of the words in the vocabulary are replaced by pseudowords), the cosine similarity remains above $0.7$, which is significant in the $300$ dimensional space. 
This provides positive support for our analysis. 
  
%First, train GloVe word embeddings using the Wikipedia corpus. Now pick two  random  words $w_1, w_2$ and replace each of their occurrences with a newly minted word $w$. Then compute an embedding for $w$ while keeping all other  embeddings unchanged. Repeating such an experiment for $100$ pairs of $w_1, w_2$ with various values of frequencies,  we find that $v_{w} \approx  \alpha_1  v_{w_1} + \alpha_2v_{w_2}$  with average $\ell_2$ error less than $6\%$ and the Pearson correlation to the formula for the coefficients in Theorem~\ref{thm:main1} is about $0.73$. 

\noindent\textbf{Test 2.}
The second test is a proxy for what would be a complete (but laborious) test of the Linearity Assertion: replicating the thought experiment while hand-labeling sense usage for many words in a corpus. The simpler proxy is as follows. For each  word $w$,  WordNet~\cite{fellbaum} lists its various senses by providing definition and example sentences for each sense. This is enough text (roughly a paragraph's worth) for our theory to allow us to represent it by a vector ---specifically, apply the SIF sentence embedding followed by the linear transformation learned as in Section~\ref{sec:ellipsoidtest}. The text embedding  for sense $s$ should approximate the ground truth vector $v_s$ for it. Then the Linearity Assertion predicts that embedding $v_w$ lies close to the subspace spanned by the sense vectors. (Note that this is a nontrivial event: in $300$ dimensions a random vector will be quite far from the subspace spanned by some $3$ other random vectors.) 
 Table~\ref{tab:closeness} checks this prediction using the polysemous words appearing in the WSI task of SemEval 2010. We tested three standard word embedding methods: GloVe, the skip-gram variant of word2vec, and SN~\cite{randomdiscourses}. The results show that the word vectors are quite close to the subspace spanned by the senses. %The SN vectors achieve the best result probably because they are computed directly using the random walk model. 

\begin{table}[!t]
	\centering
		\begin{tabular}{c| c c c}
			\hline 
			vector type & GloVe & skip-gram & SN \\
			\hline
			cosine & 0.72  & 0.73 & 0.76 \\ 
			\hline
		\end{tabular}
	\caption{The average cosine of the angles between the vectors of words and the span of vector representations of its senses. The words tested are those in the WSI task of SemEval 2010.}
	\label{tab:closeness}
\end{table}

\iffalse
We do two empirical tests of the Linearity Assertion (Theorem~\ref{thm:main1}).
 
 The first involves the classic artificial polysemous word. First, train GloVe word embeddings using the Wikipedia corpus. Now pick two  random  words $w_1, w_2$ and replace each of their occurrences with a newly minted word $w$. Then compute an embedding for $w$ while keeping all other  embeddings unchanged.
Repeating such an experiment for $100$ pairs of $w_1, w_2$ with various values of frequencies,  we find that $v_{w} \approx  \alpha_1  v_{w_1} + \alpha_2v_{w_2}$  with average $\ell_2$ error less than $6\%$ and the Pearson correlation to the formula for the coefficients in Theorem~\ref{thm:main1} is about $0.73$. 

\fi

\section{Experiments with Atoms of Discourse}  \label{sec:exp}

%\begin{table*}
	%\centering% Centers the contents of the minipage     
	%{\small
		%\begin{tabular}{|l|l|}
			%\hline  & sentence \\
			%\hline  \multirow{2}{*}{1} & The spectrum of any commutative ring with the Zariski topology (that is, the set of \\ 
			%& all prime ideals) is compact.  \\ 
			%\hline  2 & The inner 15-point ring is guarded with 8 small bumpers or posts. \\ 
			%\hline  3& Allowing a Dect phone to ring and answer calls on behalf of a nearby mobile phone. \\ 		
			%\hline   4&  The inner plastid-dividing ring is located in the inner side of the chloroplast's inner.\\ 
			%\hline  5&  \parbox[t]{0.8\textwidth}{Goya (wrestler), ring name of Mexican professional wrestler Gloria Alvarado Nava.} \\ 
			%\hline  \multirow{2}{*}{6} & The Chalk Emerald ring, containing a top-quality 37-carat emerald, in the U.S. \\ 
			%& National Museum of Natural History.\\ 
			%\hline  7 & \parbox[t]{0.8\textwidth}{Typically, elf circles were fairy rings consisting of a ring of small mushrooms.}  \\ 
			%\hline
		%\end{tabular} 
	%}   
	%\caption{Relevant fragments from top 7 sentences identified by the algorithm for the word {\em ring}. The math sense in the first sentence was missing in WordNet.}
	%\label{tab:sent_ring}
%\end{table*}

\begin{table*}[!h]
	\centering% Centers the contents of the minipage
	{\small
		\begin{tabular}{|l|l|l|l|l|l|l|}
			\hline  Atom  1978 & 825  & 231   & 616 & 1638 & 149  & 330  \\ 
			\hline 
			drowning  & instagram & stakes & membrane  & slapping & orchestra  & conferences \\ 
			%\hline 
			suicides & twitter & thoroughbred & mitochondria & pulling & philharmonic & meetings  \\ 
			%\hline 
			overdose  & facebook &guineas  & cytosol & plucking & philharmonia & seminars \\ 
			%\hline 
			murder & tumblr  & preakness & cytoplasm & squeezing & conductor  & workshops \\ 
			%\hline 
			poisoning &vimeo  & filly & membranes  & twisting & symphony & exhibitions \\ 
			%\hline 
			commits  & linkedin &  fillies & organelles & bowing & orchestras & organizes \\ 
			%\hline 
			stabbing & reddit & epsom & endoplasmic  & slamming  & toscanini & concerts \\ 
			%\hline 
			strangulation & myspace & racecourse & proteins & tossing & concertgebouw & lectures \\ 
			%\hline 
			gunshot  & tweets & sired  & vesicles & grabbing & solti & presentations \\ 
			\hline 
		\end{tabular} 
	}
	\caption{Some discourse atoms and their nearest $9$ words. By Equation~(\ref{eqn:discoursemodel}), words most likely to appear in a discourse are those nearest to it.  } 
	\label{tab:discourse}
	%\vspace{-0.1in}
\end{table*}

\begin{table*}
\centering
{
	\small
	\setlength\tabcolsep{3pt}
\begin{tabular}{|l|l|l|l|l||l|l|l|l|l|}
\hline \multicolumn{5}{|c||}{tie} &  \multicolumn{5}{|c|}{spring} \\
\hline 
trousers &season &  scoreline  & wires  & operatic  &  beginning    &    dampers    &      flower     &      creek         &  humid  \\
blouse & teams  & goalless & cables  & soprano & until      &    brakes &        flowers  &         brook   &      winters \\
 	 waistcoat & winning & equaliser  & wiring & mezzo  &months      &suspension       &flowering      &     river    &     summers  \\
skirt  & league  &  clinching~ & electrical &  contralto &  earlier      & absorbers        &fragrant           & fork     &       ppen  \\
 sleeved  &  finished & scoreless  & wire  & baritone  & year         & wheels          &lilies           & piney            &warm  \\
pants  & 	 championship & replay  & cable  & 	 coloratura  & 	 last        &  damper        &flowered     &        elk    &temperatures \\
\hline 
\end{tabular}    
}
\caption{Five discourse atoms linked to the words {\em tie} and {\em spring}. Each atom is represented by its nearest $6$ words. The algorithm often makes a mistake in the last atom (or two), as happened here.}
\label{tab:representation}
\end{table*}

%\yingyu{This section seems to overlap with or relate to practical details in Section 3, and Section 3.1. shall we merge them? }
The experiments use $300$-dimensional embeddings created using the SN objective in \cite{randomdiscourses} 
%~\eqref{eqn:PMIobj} %the model in~\cite{randomdiscourses} 
and a Wikipedia corpus of 3 billion tokens~\cite{enwiki}, and the sparse coding is solved by standard $k$-SVD algorithm~\cite{ivan2010a}. %As explained in Section~\ref{sec:method},
Experimentation showed that 
the best sparsity parameter $k$ (i.e., the maximum number of allowed senses per word) is $5$, and the number of atoms $m$ is about $2000$. 
%This hyperparameter choice is detailed below. 
%\input{figures-meta-atom/figure_sciences_atoms}
%\iffalse 
For the number of senses $k$, we tried plausible alternatives (based upon suggestions of many colleagues) that allow $k$ to vary for different words, for example to let $k$ be correlated with the word frequency. 
But a fixed choice of $k=5$ seems to produce just as good results.
 To understand why, realize that this method retains no information about the corpus except for the low dimensional word embeddings. 
Since the sparse coding tends to express a word using fairly different atoms, examining (\ref{eqn:sparsecoding}) shows that
$\sum_j \alpha^2_{w,j}$ is bounded by approximately $\|v_w\|^2_2$. 
So if too many $\alpha_{w,j}$'s are allowed to be nonzero, then some must necessarily have small coefficients, which makes the corresponding components indistinguishable from noise. In other words, raising $k$ often picks not only atoms corresponding to additional senses, but also many that don't. 
%Fixing this (possibly by going back to the co) %atoms that do not correspond to true senses. 
% Handling this imprecision issue is left for future work. 

The best number of atoms $m$ was found to be around $2000$. This was estimated by re-running the sparse coding algorithm multiple times with different random initializations, whereupon substantial overlap was found between the two bases:  a large fraction of vectors in one basis were found to have a very close vector in the other. Thus combining the bases while merging duplicates yielded a basis of about the same size. Around 100 atoms are used by a large number of words or have no close-by words. They appear semantically meaningless and are excluded by checking for this condition.\footnote{We think semantically meaningless atoms ---i.e., unexplained inner products---exist because a simple language model such as ours cannot explain all observed co-occurrences due to grammar,
stopwords, etc. It ends up needing smoothing terms. \label{foot:smooth}}

%As mentioned, each atom represents some clear and narrow \textquotedblleft topic,\textquotedblright\ and its content can be discerned by looking at the nearby words in cosine similarity. 
The content of each atom can be discerned by looking at the nearby words in cosine similarity. 
Some examples are shown in Table~\ref{tab:discourse}.
Each word is represented using at most five atoms, which usually capture distinct senses (with some noise/mistakes). The senses recovered for {\em tie} and {\em spring} are shown in Table~\ref{tab:representation}.
Similar results can be obtained by using other word embeddings like word2vec and GloVe.

%
%\noindent\textbf{Outputting relevant sentences.} Now we describe how to
%extract, for  each sense of a polysemous word, some representative sentences from the corpus.
%This would be helpful in automated construction of WordNet-like lexicons.
 %Noting that sentences in general correspond to more than one discourse atom (simply  because the number of possible things one could talk about exceeds the number of atoms), we define the {\em semantic representation} of a sentence to be the best rank-$3$ approximation (via Principal Component Analysis) to the subspace spanned by the word embeddings of its words. 
%For a given polysemous word we take its five atoms as well as atoms for its inflectional forms (past tense, plural etc., generated by~\newcite{manning-EtAl:2014:P14-5}). This yields 10 to 20 atoms for the word, whereupon each sentence is scored with respect to each atom using cosine similarity between the atom and the semantic representation of the sentence. Finally output the top few sentences with highest scores. Table~\ref{tab:sent_ring} presents the sentences found for the word {\em ring}. 
%%Due to space limitation, more examples are presented in the full version.

We also observe sparse coding procedures assign nonnegative values to most coefficients $\alpha_{w,j}$'s even if they are left unrestricted.
% and nevertheless almost all positive even though they are unconstrained, 
Probably this is because the appearances of a word are best explained by what discourse {\em is} being used to generate it, rather than what discourses are {\em not} being used.

\noindent\textbf{Relationship to Topic Models.} \label{sec:hierarchy}
Atoms of discourse may be reminiscent of results from other automated methods for obtaining a thematic understanding of text, such as topic modeling, described in the survey by~\newcite{blei2012probabilistic}. This is not surprising since the model~(\ref{eqn:discoursemodel}) used to compute the embeddings is related to a log-linear topic model by~\newcite{mnih2007three}. 
 However, the discourses here are computed via sparse coding on word embeddings, which can be seen as a linear algebraic alternative, resulting in fairly fine-grained topics.
 % very distinct from topic modeling. The 
 Atoms are also reminiscent of coherent \textquotedblleft word clusters\textquotedblright\ detected in the past using Brown clustering, or even sparse coding~\cite{murphytm12learning}. The novelty in this paper is a clear interpretation of the sparse coding results as atoms of discourse, as well as its use to capture different word senses.

\iffalse 
It's possible to further cluster discourse atoms to get meta atoms, which are highly reminiscent of past results from hierarchical topic models~\cite{griffiths2004hierarchical}.  For instance, we found  discourse atoms for {\em jazz, rock,} {\em classical} and {\em country}. These
are  more related to each other than to atoms about, say, {\em mathematics}. Again, model (\ref{eqn:discoursemodel}) suggests that similar discourse atoms should have higher inner products to each other, and thus sparse coding should be able to identify these similarities and create meta-discourse atoms such as {\em music}.
 
The same hyperparameter selection approach as above leads to sparsity $2$ and basis size $200$ for sparse coding on discourse atoms to find meta-discourse atoms. 
Figure~\ref{fig:metaatom_sciences} shows an example for scientific fields.  A discourse about an interdisciplinary science like {\em biochemistry} turns out to be approximately linear combinations of two meta-discourses of {\em biology} and {\em chemistry}. Due to space limitation, more examples are presented in the supplementary.\footnote{This will be released upon publication of the paper.} 

\fi

\section{Testing WSI in Applications} \label{sec:test}  

While the main result of the paper is to reveal the linear algebraic structure of word senses within existing embeddings, it is desirable to verify that this view can yield results competitive with earlier sense embedding approaches. We report some tests below. We find that common word embeddings perform similarly with our method; for concreteness we use induced embeddings described in Section~\ref{sec:ellipsoidtest}.
\iffalse Start with GloVe embeddings, and use them to compute  \textquotedblleft induced embeddings\textquotedblright for all words as follows. Randomly sample 1 million sentences from Wikipedia, compute the average discourse vector for each word, and learn a linear transformation according to (\ref{eqn:2}), and used the transformed average discourse vectors as the induced word vectors. \fi
They are evaluated in three tasks: word sense induction task in SemEval 2010~\cite{manandhar2010semeval}, word similarity in context~\cite{huang2012improving}, and a new task we called police lineup test. 
The results are compared to those of existing embedding based approaches reported in related work~\cite{huang2012improving,neelakantan2014efficient,mu2016geometry}. %\footnote{Note that~\cite{mu2016geometry} is a followup to earlier arxiv versions of the current paper.}. % The current version has a somewhat better performance.}.
\subsection{Word Sense Induction} \label{sec:semeval}

In the WSI task in SemEval 2010, the algorithm is given a polysemous word and about 40 pieces of texts, each using it according to a single sense. The algorithm has to cluster the pieces of text so that those with the same sense are in the same cluster. The evaluation criteria are F-score~\cite{artiles2009role} and V-Measure~\cite{rosenberg2007v}. The F-score tends to be higher with a smaller number of clusters and the V-Measure tends to be higher with a larger number of clusters, and fair evaluation requires reporting both.

Given a word and its example texts, our algorithm uses a Bayesian analysis dictated by our theory to compute a vector $u_c$ for the word in each context $c$ and and then applies $k$-means on these vectors, with the small twist that sense vectors are assigned to nearest centers based on inner products rather than Euclidean distances. Table~\ref{tab:semeval} shows the results.

\noindent\textbf{Computing vector $u_c$.}
For word $w$ we start by computing its expansion in terms of atoms of discourse (see (\ref{eqn:sparsecoding_obj}) in Section~\ref{sec:method}). In an ideal world the nonzero coefficients would exactly capture its senses, and each text containing $w$ would match to one of these nonzero coefficients. In the real world such deterministic success is elusive and one must reason using Bayes' rule.

For each atom $a$, word $w$ and text $c$ there is a joint distribution $p(w, a, c)$ describing the event that atom $a$ is the sense being used when word $w$ was used in text $c$. Assuming that $p(w,c|a) = p(w|a)p(c|a)$ (similar to Eqn~(\ref{eqn:discoursemodel})), the posterior distribution is:
\begin{align} \label{eqn:sense_prob}
   p(a | c, w) \propto  p( a | w ) p( a | c) / p(a).
\end{align}
We approximate $p(a | w)$ using Theorem~\ref{thm:main1}, which suggests that the coefficients in the expansion of 
$v_w$ with respect to atoms of discourse scale according to probabilities of usage. (This assertion involves ignoring the low-order terms involving the logarithm in the theorem statement.) Also, by the random walk model, $p(a|c)$ can be approximated by $\exp(\langle v_a, v_c \rangle)$ where $v_c$ is the SIF embedding of the context. Finally, since $p(a) = \mathbf{E}_c [p(a|c)]$, it can be empirically estimated by randomly sampling $c$.

\begin{table}[!tb]
	\centering
		\begin{tabular}{c| c c c}
		\hline
		Method & V-Measure & F-Score  \\
		  \hline 
			%Duluth-WSI & 9.0 & 41.1  \\
			%UoY &  15.7 & 49.8  \\
			%KCDC-PC & 7.5 & 55.5  \\
			%\hline 
			\cite{huang2012improving} & 10.60 & 38.05\\
			\cite{neelakantan2014efficient} &  9.00 & 47.26 \\
      %\cite{neelakantan2014efficient}, MSSG.300D.6K.key & 6.90 &  48.43  \\
			\cite{mu2016geometry},  $k=2$ & 7.30 & \textbf{57.14} \\
			\cite{mu2016geometry},  $k=5$ & \textbf{14.50} & 44.07 \\	
		\hline
			ours,  $k=2$ & 6.1 & \textbf{58.55}\\
			ours,  $k=3$ & 7.4 & 55.75 \\
			ours,  $k=4$ & 9.9 & 51.85 \\
			ours,  $k=5$ & \textbf{11.5} & 46.38 
			\\
		\hline
		\end{tabular}
	\caption{Performance of different vectors in the WSI task of SemEval 2010. The parameter $k$ is the number of clusters used in the methods. Rows are divided into two blocks, the first of which shows the results of the competitors, and the second shows those of our algorithm. Best results in each block are in boldface.}
	\label{tab:semeval}
\end{table}

The posterior $p(a | c, w)$ can be seen as a soft decoding of text $c$ to atom $a$. If texts $c_1, c_2$ both contain $w$, and they were hard decoded to atoms $a_1, a_2$ respectively then their similarity would be $\langle v_{a_1}, v_{a_2} \rangle$. With our soft decoding, the similarity can be defined by taking the expectation over the full posterior:
\begin{align} \label{eqn:wc_sim}
  &\quad ~ \text{similarity}(c_1, c_2) \nonumber\\
	& =  \mathbf{E}_{a_i  \sim   p(a|c_i, w), i\in\{1,2\}} \langle v_{a_1}, v_{a_2} \rangle, \\
	 &  = \left\langle\sum_{a_1} p(a_1 | c_1, w) v_{a_1}, \sum_{a_2} p(a_2 | c_2, w) v_{a_2} \right\rangle.\nonumber
\end{align}
At a high level this is analogous to the Bayesian polysemy model of \newcite{reisingermooney} and \newcite{brody2009bayesian}, except that they introduced separate embeddings for each sense cluster, while here we are working with structure already existing inside word embeddings.
 
The last equation suggests defining the vector $u_c$ for the word $w$ in the context $c$ as
\begin{align} \label{eqn:context_vec}
   u_c = \sum_{a} p(a | c, w) v_a,
\end{align}
which allows the similarity of the word in the two contexts to be expressed via their inner product.

\noindent\textbf{Results.} The results are reported in Table~\ref{tab:semeval}. %\yingyu{Need to add some results from SemEval submissions}
Our approach outperforms the results by \newcite{huang2012improving} and \newcite{neelakantan2014efficient}. 
When compared to~\newcite{mu2016geometry}, for the case with 2 centers, we achieved better V-measure but lower F-score, while for 5 centers, we achieved lower V-measure but better F-score.

\subsection{Word Similarity in Context}
The dataset consists of around 2000 pairs of words, along with the contexts the words occur in and the ground-truth similarity scores. The evaluation criterion is the correlation between the ground-truth scores and the predicted ones.
Our method computes the estimated sense vectors and then the similarity as in Section~\ref{sec:semeval}. 
We compare to the baselines that simply use the cosine similarity of the GloVe/skip-gram vectors, and also to the results of several existing sense embedding methods. 

\noindent\textbf{Results.} 
Table~\ref{tab:scws} shows that our result is better than those of the baselines and \newcite{mu2016geometry}, but slightly worse than that of \newcite{huang2012improving}. Note that \newcite{huang2012improving} retrained the vectors for the senses on the corpus, while our method depends only on senses extracted from the off-the-shelf vectors. After all, our goal is to show word senses already reside within off-the-shelf word vectors. %It is possible to extend our method to utilize the corpus for better performance. 
%Our algorithm using the average discourse vectors without linear transformation leads to a significantly worse result, which supports our Gaussian Walk model.

%Note that simply whitening the transformed average discourse vectors already leads to significant improvement, achieving performance comparable to \cite{huang2012improving}. And incorporating the context using our algorithm only leads to a slight improve ment. 

%\begin{table}[!t]
	%\centering
		%\begin{tabular}{c|c}
		%\hline		
  %Method   & Spearman coefficient \\
	%\hline
  %GloVe       &     0.573 \\
	%skip-gram   &     0.622  \\
  %%transformed average discourse vectors  &    0.6248 \\
	%%whitened transformed average discourse vectors & 0.6516 \\
	%\cite{huang2012improving} &  \textbf{0.657} \\
	%\cite{neelakantan2014efficient} & 0.567 \\
	%\cite{mu2016geometry} & 0.637 \\
	%\hline
		%ours, induced embeddings &    \textbf{0.652}\\
		%ours, no transformation    & 0.397 \\
		%ours, GloVe  &    0.584 \\
		%\hline
		%\end{tabular}
	%\caption{The results for different methods in the SCWS task. The rows are divided into blocks. The first blocks are for the competitors. The second is for our algorithm using different word vectors: the induced embeddings, the average discourse vectors ($u_w$ in Section~\ref{sec:ellipsoidtest}), and GloVe vectors. The best result in each block is in boldface. }
	%\label{tab:scws}
%\end{table}

\begin{table}[!t]
	\centering
		\begin{tabular}{c|c}
		\hline		
  Method   & Spearman coefficient \\
	\hline
  GloVe       &     0.573 \\
	skip-gram   &     0.622  \\
	\cite{huang2012improving} &  \textbf{0.657} \\
	\cite{neelakantan2014efficient} & 0.567 \\
	\cite{mu2016geometry} & 0.637 \\
	\hline
		ours &    0.652\\
		\hline
		\end{tabular}
	\caption{The results for different methods in the task of word similarity in context. The best result is in boldface. Our result is close to the best.}
	\label{tab:scws}
\end{table}

\subsection{Police Lineup}

Evaluating WSI systems can run into well-known difficulties, as reflected in the changing metrics over the years~\cite{navigli2013semeval}. %Some metrics involve a custom similarity score based upon WordNet that is hard to interpret and may not be available for other languages. 
Inspired by word-intrusion tests for topic coherence~\cite{chang2009reading}, we proposed a new simple test, which has the advantages of being easy to understand, and capable of being administered to humans. 

The testbed
%, available in the supplementary,\footnote{The testbed will be made public upon publication of the paper.} 
uses 200 polysemous words and their 704 senses according to WordNet. Each sense is represented by 8 related words, which were collected from WordNet and online dictionaries by college students, who were told to identify most relevant other words occurring in the online definitions of this word sense as well as in the accompanying illustrative sentences. These are considered as ground truth representation of the word sense. These $8$ words are typically not synonyms. For example, for the {\em tool/weapon} sense of {\em axe} they were  ``handle, harvest, cutting,  split, tool, wood, battle, chop.''

\begin{table}[!t]
	\centering
	{\scriptsize
		\begin{tabular}{c|c|l}
		\hline		
  word  &  \multicolumn{2}{c}{senses} \\
	\hline
	\multirow{6}{*}{bat} 
	       & 1& \textbf{navigate	nocturnal	mouse	wing	cave	sonic	fly	dark} \\
	       & 2& \textbf{used	hitting	ball	game	match	cricket	play	baseball} \\
         & 3&  \textbf{wink	briefly	shut	eyes	wink	bate	quickly	action} \\
				& 4& whereby legal court law lawyer suit bill judge \\
				& 5& loose ends two loops shoelaces tie rope string \\
				& 6 & horny projecting bird oral nest horn hard food \\ 
		\hline
		\end{tabular}
		}
	\caption{An example of the police lineup test with $n=6$. The algorithm (or human subject) is given the polysemous word ``bat'' and $n=6$ senses each of which is represented as a list of words, and is asked to identify the true senses belonging to ``bat'' (highlighted in boldface for demonstration).}
	\label{tab:policelineup_example}
\end{table}

\begin{algorithm}[!th]
\caption{Our method for the police lineup test} \label{alg:test}
\begin{algorithmic}[1]
\REQUIRE{\small Word $w$, list $S$ of senses (each has $8$ words)}
\ENSURE{\small $t$ senses out of $S$}
\STATE  \small Heuristically find inflectional forms of $w$. \normalsize
\STATE \small Find $5$ atoms for $w$ and each inflectional form. Let $U$ denote the union of all these atoms.
\STATE \small Initialize the set of candidate senses $C_w \leftarrow \emptyset$,
and the score for each sense $L$ to $\text{score}(L) \leftarrow -\infty$
\normalsize
\FOR{\small each atom $a \in U$\normalsize}
\STATE \small  Rank senses $L \in S$ by \\
$\hspace{5mm}\text{score}(a, L) \!= \!s(a, L) \!- \! s_A^L + s(w, L)  - s_V^L$\normalsize 
\STATE \small 
Add the two senses $L$ with highest $\text{score}(a, L)$ to $C_w$, and update their scores\\
$\hspace{5mm}\text{score}(L) \leftarrow \max\{\text{score}(L), \text{score}(a, L) \}$
\normalsize
\ENDFOR
\STATE \small Return the $t$ senses $L \in C_s$ with highest $\text{score}(L)$\normalsize
\end{algorithmic}
\end{algorithm}

\begin{figure*}[!th]
	\centering
		\begin{tabular}{cc c}
		\includegraphics[width=0.85\columnwidth]{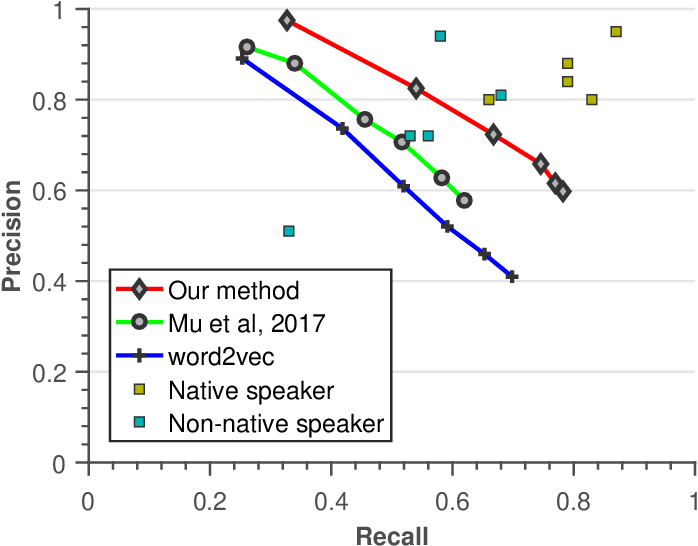} %{figure/all_m20_s2}
		& \qquad\qquad&
		\includegraphics[width=0.85\columnwidth]{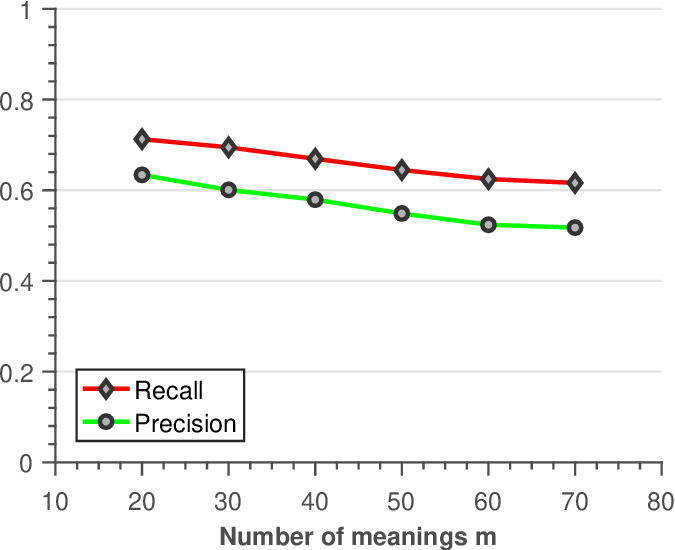} \\ %{figure/r4_s2}\\
		\textbf{A} && \textbf{B}
		\end{tabular}
		\vspace{-0.1in}
		\caption{Precision and recall in the police lineup test. (\textbf{A}) For each polysemous word, a set of $n = 20$ senses containing the ground truth senses of the word are presented. Human subjects are told that on average each word has 3.5 senses and were asked to choose the senses they thought were true. The algorithms select $t$ senses for $t=1,2,\dots, 6$. For each $t$, each algorithm was run 5 times (standard deviations over the runs are too small to plot).  (\textbf{B}) The performance of our method for $t=4$ and $n=20,30,\dots,70$.	
		}
	\label{fig:polysemytest}
\end{figure*}
%\begin{figure}[!th]
%	\centering
%
%		\includegraphics[width=0.7\columnwidth]{figure/all_m20_s2}
%
%		\includegraphics[width=0.6\columnwidth]{figure/r4_s2}\\
%	\caption{Precision and recall in the polysemy test. (\textbf{A}) For each polysemous word, a set of $n = 20$ senses containing the ground truth senses of the word are presented. Human subjects are told that on average each word has 3.5 senses and were asked to choose the senses they thought were true. The algorithms select $t$ senses for $t=1,2,\dots, 6$. For each $t$, each algorithm was run 5 times (standard deviations over the runs are $<0.02$ and thus not plotted).  (\textbf{B}) The performance of our method for $t=4$ and $n=20,30,\dots,70$.	
%	}
%	\label{fig:polysemytest}
%\end{figure}

The quantitative test is called {\em police lineup}.
First, randomly pick one of these 200 polysemous words. Second, pick the true senses for the word and then add randomly picked senses from other words so that there are $n$ senses in total, where each sense is represented by 8 related words as mentioned. Finally, the algorithm (or human) is given the polysemous word and a set of $n$ senses, and has to identify the true senses in this set. Table~\ref{tab:policelineup_example} gives an example. \yingyu{reviewer B asks for some details. }

Our method (Algorithm~\ref{alg:test}) uses the similarities between any word (or atom) $x$ and a set of words $Y$, defined as
$s(x, Y) = \inner{v_x, v_Y}$ where $v_Y$ is the SIF embedding of $Y$. 
It also uses the average similarities: 
%\small
$$
 s_A^Y = \frac{\sum_{a\in A} s(a, Y)}{|A|},  ~~s_V^Y = \frac{\sum_{w\in V} s(w, Y)}{|V|}
$$
%\normalsize
where $A$ are all the atoms, and $V$ are all the words.
We note two important practical details.
First, while we have been using atoms of discourse as a proxy for word sense, these are too coarse-grained: the total number of senses (e.g., WordNet synsets) is far greater than $2000$. Thus the score($\cdot$) function uses both the atom and the word vector. Second, some words are more popular than the others---i.e., have large components along many atoms and words---which seems to be an instance of the
smoothing phenomenon alluded to in Footnote~\ref{foot:smooth}. %Section~\ref{sec:pastmodels}. % {sec:exp} . 
The penalty terms $s_A^L$ and $s_V^L$ lower the scores of senses $L$ containing such words. 
Finally, our algorithm returns $t$ senses where $t$ can be varied.

%The precision (fraction of recovered senses that are correct) and recall (fraction of ground truth senses recovered) 
\noindent\textbf{Results.} The precision and recall
for different $n$ and $t$ (number of senses the algorithm returns) are presented in Figure~\ref{fig:polysemytest}. Our algorithm outperforms the two selected competitors. For $n=20$ and $t=4$, our algorithm succeeds with precision  $65\%$ and recall $75\%$, and performance remains reasonable for $n=50$.
Giving the same test to humans\footnote{Human subjects are graduate students from science or engineering majors at major U.S. universities. Non-native speakers have 7 to 10 years of English language use/learning.}
for $n=20$  (see the left figure) suggests that our method performs similarly to non-native speakers.

Other word embeddings can also be used in the test and  achieved slightly lower performance. For $n =20$ and $t=4$, the precision/recall are lower by the following amounts: 
{GloVe} $2.3\%/5.76\%$, 
%{word2vec}\footnote{The variant CBOW in the {word2vec} family was used.} $4.3\%/9.9\%$, 
{NNSE} (matrix factorization on PMI to rank 300 by \newcite{murphytm12learning}) $25\%/28\%$.

\section{Conclusions} \label{sec:conclusion}
Different senses of polysemous words have been shown to lie in linear superposition inside standard word embeddings like  word2vec and GloVe. This has also been shown theoretically building upon previous generative models, and empirical tests of this theory were presented. A priori, one imagines that showing such theoretical results about the inner structure of modern word embeddings would be hopeless since they are solutions to complicated nonconvex optimization.

A new  WSI method is also proposed based upon these insights that uses only the word embeddings and  sparse coding, and shown to provide very competitive performance on some WSI benchmarks. 
%Currently the WSI seems to do better with nouns than other parts of speech, and improving this is left for future work. 
One novel aspect of our approach is that the word senses are interrelated using one of about $2000$ discourse vectors that give a succinct description of which other words appear in the neighborhood with that sense.
% This makes the method potentially more useful for other tasks in NLP, for example, automated creation of WordNets in other languages. 
Our method based on sparse coding can be seen as a linear algebraic analog of the clustering approaches, and also gives fine-grained thematic structure reminiscent of topic models. 

A novel police lineup test was also proposed 
for testing such WSI methods, where the algorithm is given a word $w$ and word clusters, some of which belong to senses of $w$ and the others are distractors belonging to senses of other words.  The algorithm has to identify the ones belonging to $w$.
We conjecture this police lineup test with distractors will challenge some existing WSI methods, whereas our method was found to achieve performance similar to non-native speakers.

%The method can be more useful and accurate in a semi-automated mode, since human helpers are better at {\em recognizing} a presented word sense than at coming up with a complete list. 
%Thus instead of listing only 5 senses per word as given by the sparse coding, the algorithm can ask a %human helper to examine the list of $20$ closest discourse atoms (and sample sentences as in Section~\ref{sec:exp}) which usually gives high recall  for senses  missing in the top $5$.  

\iffalse 
As mentioned in Section~\ref{sec:linear}, the use of logarithm in these PMI-like embeddings   seems key to the success of our approach, because the logarithm allows less frequent senses to have a higher weight in the linear superposition relative to their frequency. Without the logarithm the
less frequent sense would be hard to distinguish from noise.   This may have relevance in neuroscience, where word embeddings have been used in fMRI studies to map what the subject {\em is thinking about}~\cite{mitchelletal}. The words in that study were largely monosemous, and the more nuanced word embeddings and discourse vectors introduced in this work may be useful in further explorations, especially since sparse coding as well as logarithms are thought to be neurally plausible.
\fi

\section*{Acknowledgements}

We thank the reviewers and the Action Editors of TACL for helpful feedbacks and thank the editors for granting a special relaxation of the page limit for our paper. 
This work was supported in part by NSF grants %CCF-08327, 
%CCF-1117309, CCF-1302518, 
CCF-1527371, DMS-1317308, Simons Investigator Award, Simons Collaboration Grant,
and ONR-N00014-16-1-2329. Tengyu Ma was additionally supported by the Simons Award in Theoretical Computer Science and by the IBM Ph.D.\ Fellowship.

\bibliographystyle{acl2012}
\bibliography{polysemy}

\end{document}